\documentclass[a4paper,twoside]{article}

\usepackage{epsfig}
\usepackage{subfigure}
\usepackage{calc}
\usepackage{amssymb}
\usepackage{amstext}
\usepackage{amsmath}
\usepackage{amsthm}
\usepackage{multicol}
\usepackage{pslatex}
\usepackage{apalike}

\usepackage{csquotes}
\usepackage[capitalize]{cleveref}
\usepackage{pgfplots}
\usepackage[]{algorithmicx}
\usepackage[]{algpseudocode}
\usepackage{float}
\usepackage{multirow}
\usepackage{pbox}
\usepackage[export]{adjustbox}
\usepackage{hyphenat}
\usepackage{array}
\usepackage{url}
\usepackage{tabularx}
\usepackage{xspace}

\usepackage{SCITEPRESS}     

\floatstyle{plaintop}
\newfloat{algorithm}{h}{lop}
\floatname{algorithm}{Algorithm}

\makeatletter
\algrenewcommand\ALG@beginalgorithmic{\ttfamily\small}
\makeatother

\newcommand{\etal}{\emph{et al.}\xspace}
\newcommand{\eg}{\emph{e.g.,}\xspace}
\newcommand{\ie}{\emph{i.e.,}\xspace}
\newcommand{\wrt}{\emph{w.r.t.}\xspace}

\newcolumntype{Y}{>{\centering\arraybackslash}X}
\newcolumntype{L}[1]{>{\raggedright\arraybackslash}p{#1}}
\pgfplotsset{
	legend entry/.initial=,
	every axis plot post/.code={%
		\pgfkeysgetvalue{/pgfplots/legend entry}\tempValue
		\ifx\tempValue\empty
		\else
		\expandafter\addlegendentry\expandafter{\tempValue}%
		\fi
	},
}

\newenvironment{experiplotex}[8]
{
	\begin{tikzpicture}%
	\begin{axis}[style={font=\small},scale only axis, clip bounding box=upper bound, enlarge x limits=0.05, enlarge y limits=0.1, title={#1},width=#6, height=#7,cycle list name=color list, legend style={#8}, xlabel={#4},ylabel={#5},
	y label style={at={(axis description cs:0.0,.5)},anchor=north},
	y tick label style={/pgf/number format/.cd, fixed, fixed zerofill, precision=1, /tikz/.cd}]%
}
{%
	\end{axis}
	\end{tikzpicture}%
}

\begin{document}

\newif\ifVISAPP
\VISAPPfalse

\title{Active Learning for Deep Object Detection}

\ifVISAPP
\author{Anonymous VISAPP Submission \#6}
\else
\author{\authorname{Clemens-Alexander Brust\sup{1}, Christoph K{\"a}ding\sup{1,2} and Joachim Denzler\sup{1,2}}
\affiliation{\sup{1}Computer Vision Group, Friedrich Schiller University Jena, Germany}
\affiliation{\sup{2}Michael Stifel Center Jena, Germany}
\email{\{f\_author, s\_author\}@uni-jena.de}
}
\fi


\abstract{The great success that deep models have achieved in the past is mainly owed to large amounts of labeled training data. However, the acquisition of labeled data for new tasks aside from existing benchmarks is both challenging and costly. Active learning can make the process of labeling new data more efficient by selecting unlabeled samples which, when labeled, are expected to improve the model the most.
In this paper, we combine a novel method of active learning for object detection with an incremental learning scheme \cite{Kaeding2016FDN} to enable continuous exploration of new unlabeled datasets. We propose a set of uncertainty-based active learning metrics suitable for most object detectors. Furthermore, we present an approach to leverage class imbalances during sample selection. All methods are evaluated systematically in a continuous exploration context on the PASCAL VOC 2012 dataset \cite{Everingham2010VOC}.}

\onecolumn \maketitle \normalsize \vfill


\begin{figure*}[tb]
    \centering
    \includegraphics[width=1.0\textwidth]{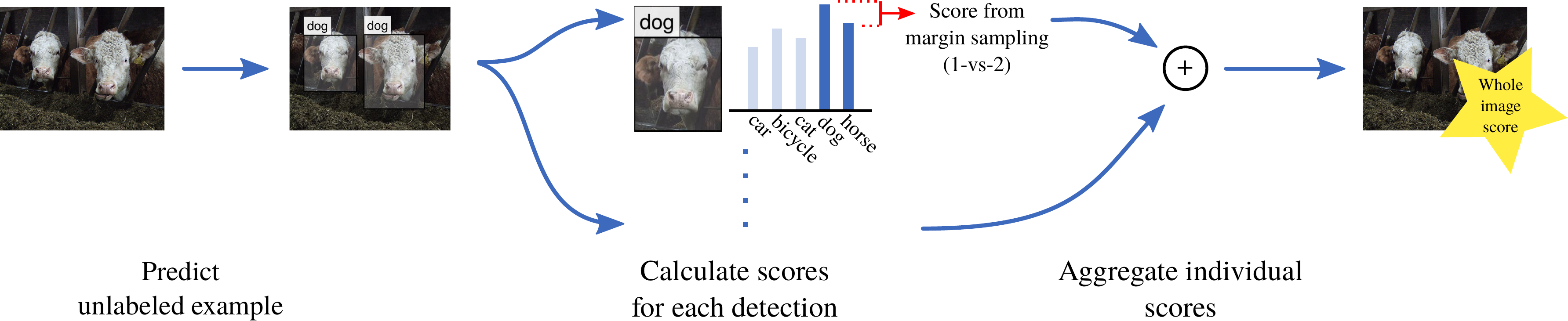}
    \vspace{0pt}
    \caption{
        Our proposed system for continuous exploration scenarios.
        Unlabeled images are evaluated by an deep object detection method.
        The margins of predictions (\ie absolute difference of highest and second-highest class score)
        are aggregated to identify valuable instances by combining scores of individual detections.}
    \label{fig:teaser}
\end{figure*}

\section{Introduction}
\label{sec:intro}
Labeled training data is highly valuable and the basic requirement of supervised learning.
Active learning aims to expedite the process of acquiring new labeled data, ordering unlabeled samples by the expected value from annotating them.
In this paper, we propose novel active learning methods for object detection.
Our main contributions are
\emph{(i)} an incremental learning scheme for deep object detectors without catastrophic forgetting based on \cite{Kaeding2016FDN},
\emph{(ii)} active learning metrics for detection derived from uncertainty estimates and
\emph{(iii)} an approach to leverage selection imbalances for active learning.

While active learning is widely studied in classification tasks \cite{Grauman2016Crowd,Settles2009ALS}, it has received much less attention in the domain of deep object detection.
In this work, we propose methods that can be used with any object detector that predicts a class probability distribution per object proposal.
Scores from individual detections are aggregated into a score for the whole image (see \cref{fig:teaser}).
All methods rely on the intuition that model uncertainty and valuable samples are likely to co-occur \cite{Settles2009ALS}.
Furthermore, we show how the balanced selection of new samples can improve the resulting performance of an incrementally learned system.

In continuous exploration application scenarios, \eg in camera streams, new data becomes available over time or the distribution underlying the problem changes itself.
We simulate such an environment using splits of the PASCAL VOC 2012 \cite{Everingham2010VOC} dataset.
With our proposed framework,
a deep object detection system can be trained in an incremental manner while the proposed aggregation schemes enable selection of valuable data for annotation.
In consequence, a deep object detector can explore unknown data and adapt itself involving minimal human supervision. This combination results in a complete system enabling continuously changing scenarios.


\subsection{Related Work}
\paragraph{Object Detection using CNNs}
An important contribution to object detection based on deep learning
is R-CNN \cite{Girshick2013RCNN}. It delivers a considerable improvement over
previously published sliding window-based approaches. R-CNN employs selective search \cite{uijlings2013selective},
an unsupervised method
to generate region proposals. A pre-trained CNN performs feature extraction.
Linear SVMs (one per class) are used to score the extracted features and a
threshold is applied to filter the large number of proposed regions.
Fast R-CNN~\cite{Girshick2015FastRCNN} and Faster R-CNN~\cite{Ren2015FasterRCNN} offer further improvements in speed and accuracy.
Later on, R-CNN is combined with feature pyramids to enable efficient multi-scale detections \cite{lin2017feature}.
YOLO \cite{Redmon2015YOLO} is a more recent deep
learning-based object detector. Instead of using a CNN as a black box feature extractor, it is trained in an end\hyp{}to\hyp{}end
fashion. All detections are inferred in a single pass (hence the name \enquote{You
Only Look Once}) while detection and classification are capable of independent operation.
YOLOv2 \cite{Redmon2016YOLOv2} and YOLOv3 \cite{redmon2018yolov3} improve upon the
original YOLO in several aspects.
These include among others different network architectures, different priors for bounding boxes and considering multiple scales during training and detection.
SSD \cite{Liu2015SSD} is a single-pass approach comparable to YOLO
introducing improvements like assumptions about the aspect ratio distribution of bounding boxes as well as predictions on different scales.
As a result of a series of improvements, it is both faster and more accurate than the original YOLO.
DSSD \cite{fu2017dssd} further improves upon SSD in focusing more on context with the help of deconvolutional layers.

\paragraph{Active Learning for Object Detection}
The authors of \cite{abramson2006active} propose an active learning system for pedestrian detection in videos taken by a camera mounted on the front of a moving car.
Their detection method is based on \mbox{AdaBoost} while sampling of unlabeled instances is realized by hand-tuned thresholding of detections.
Object detection using generalized Hough transform in combination with randomized decision trees, called Hough forests, is presented in \cite{yao2012interactive}.
Here, costs are estimated for annotations, and instances with highest costs are selected for labeling.
This follows the intuition that those examples are most likely to be difficult and therefore considered most valuable.
Another active learning approach for satellite images using sliding windows in combination with an SVM classifier and margin sampling is proposed in \cite{bietti2012active}.
The combination of active learning for object detection with crowd sourcing is presented in \cite{vijayanarasimhan2014large}.
A part-based detector for SVM classifiers in combination with hashing is proposed for use in large-scale settings.
Active learning is realized by selecting the most uncertain instances for labeling.
In \cite{roy2016active}, object detection is interpreted as a structured prediction problem using a version space approach in the so called ``difference of features'' space.
The authors propose different margin sampling approaches estimating the future margin of an SVM classifier.

Like our proposed approach, most related methods presented above rely on uncertainty indicators like least confidence or 1\hyp{}vs\hyp{}2.
However, they are designed for a specific type of object detection and therefore can not be applied to deep object detection methods in general whereas our method can.
Additionally, our method does not propose single objects to the human annotator. It presents whole images at once and requests labels for every object.

\paragraph{Active Learning for Deep Architectures}
In \cite{wang2014new} and \cite{Wang2016CEAL},
uncertainty-based active learning criteria for deep models are proposed.
The authors offer several metrics to estimate model uncertainty, including least confidence, margin or entropy sampling.
Wang \etal additionally describe a self-taught learning scheme,
where the model's prediction is used as a label for further training if uncertainty is below a threshold.
Another type of margin sampling is presented in \cite{stark2015captcha}.
The authors propose querying samples according to the quotient of the highest and second-highest class probability.
The visual detection of defects using a ResNet is presented in \cite{feng2017dal}.
The authors propose two methods: uncertainty sampling (\ie defect probability of {0.5}) and positive sampling (\ie selecting every positive sample since they are very rare)
for querying unlabeled instances as model update after labeling.
Another work which presents uncertainty sampling is \cite{liu2017active}.
In addition, a query by committee strategy as well as active learning involving weighted incremental dictionary learning for active learning are proposed.
In the work of \cite{gal2017deep}, several uncertainty-related measures for active learning are presented.
Since they use Bayesian CNNs, they can make use of the probabilistic output and employ methods like variance sampling, entropy sampling or maximizing mutual information.
Finally, the authors of \cite{beluch2018power} show that ensamble-based uncertainty measures are able to perform best in their evaluation.
All of the works introduced above are tailored to active learning in classification scenarios.
Most of them rely on model uncertainty, similar to our applied selection criteria.

Besides estimating the uncertainty of the model, further retraining-based approaches are maximizing the expected model change \cite{huang2016active}
or the expected model output change \cite{Kaeding2016AEMOC} that unlabeled samples would cause after labeling.
Since each bounding box inside an image has to be evaluated according its active learning score,
both measures would be impractical in terms of runtime without further modifications.
A more complete overview of general active learning strategies can be found in \cite{Grauman2016Crowd,Settles2009ALS}.

\section{Prerequisite: Active Learning}
\label{sec:actobjdet}
In active learning, a value or metric $v(x)$ is assigned to any unlabeled example $x$ to determine
its possible contribution to model improvement. The current model's output
can be used to estimate a value,
as can statistical properties of the example itself. A high
$v(x)$ means that the example should be preferred during selection
because of its estimated value for the current model.

In the following section, we propose a method to adapt an active learning metric
for classification to object detection using an aggregation process. This method
is applicable to any object detector whose output contains class scores for each
detected object.

\paragraph{Classification}
For classification, the model output for a given example $x$ is an estimated distribution
of class scores $\hat{p}(c|x)$ over classes $K$. This
distribution can be analyzed to determine whether the model made an uncertain prediction,
a good indicator of a valuable example.
Different measures of uncertainty are a common choice for selection, \eg
\cite{ertekin2007learning,fu2015batch,hoi2006large,jain2009active,kapoor2010gaussian,Kaeding2016WALI,tong2001support,beluch2018power}.

For example, if the difference between the two highest class scores is very low, the example
may be located close to a decision boundary. In this case, it can be used to
refine the decision boundary and is therefore valuable. The metric is defined
using the highest scoring classes $c_1$ and $c_2$:
\begin{equation}
 v_{1vs2}(x)~=~\big(1- (\underset{c_1 \in K}{\textrm{max}}\,\hat{p}(c_1|x) -
\underset{c_2 \in K\setminus c_1}{\textrm{max}}\,\hat{p}(c_2|x))\big)^2\enspace.
\end{equation}

This procedure is known as \emph{1\hyp{}vs\hyp{}2} or \emph{margin sampling} \cite{Settles2009ALS}.
We use 1\hyp{}vs\hyp{}2 as part of our methods since its operation is intuitive and it can produce better estimates than
\eg least confidence approaches \cite{Kaeding2016AEMOC}.
However, our proposed aggregation method could be applied to any other active learning measure.

\section{Active Learning for Deep Object Detection}
Using a classification metric on a single detection is straightforward, if class
scores are available.
Though, aggregating metrics of individual detections
for a complete image can be done in many different ways. In the section below,
we propose simple and efficient aggregation strategies. Afterwards,
we discuss the problem of class imbalance in datasets.

\subsection{Aggregation of Detection Metrics}
\label{sec:general}
Possible aggregations include calculating the sum, the average or the maximum
over all detections. However, for some aggregations, it is not clear how an image without
any detections should be handled.

\paragraph{Sum}
A straightforward method of aggregation is the sum.
Intuitively, this method prefers images with lots of uncertain detections in them.
When aggregating detections using a sum, empty examples should be valued zero.
It is the neutral element of addition, making it a reasonable value for an empty
sum. A low valuation effectively
delays the selection of empty examples until there are either no better examples
left or the model has improved enough to actually produce detections on them.
The value of a single example $x$ can be calculated from the detections
$D$ in the following way:
\begin{equation}
    v_{Sum}(x)~=~\sum_{i \in D} v_{1vs2}(x_i) \enspace.
\end{equation}

\paragraph{Average}
Another possibility is averaging each detection's scores.
The average is not sensitive to the number of detections, which may make scores
more comparable between images.
If a sample does not contain any detections, it will be assigned a zero score.  This is an arbitrary
rule because there is no true neutral element \wrt averages.
However, we choose zero to keep the behavior
in line with the other metrics:
\begin{equation}
    v_{Avg}(x)~=~\frac{1}{|D|}\sum_{i \in D} v_{1vs2}(x_i) \enspace.
\end{equation}

\paragraph{Maximum}
Finally, individual detection scores can be aggregated by calculating
the maximum. This can result in a substantial information loss. However,
it may also prove beneficial because of increased robustness to noise
from many detections.
For the maximum aggregation, a zero score for empty examples is valid.
The maximum is not affected by zero valued detections,
because no single detection's score can be lower than zero:
\begin{equation}
    v_{Max}(x)~=~\underset{i \in D}{\textrm{max}}~v_{1vs2}(x_i) \enspace.
\end{equation}

\subsection{Handling Selection Imbalances}
\label{sec:classwise}
Class imbalances can lead to worse results for classes underrepresented in the training
set. In a continuous learning scenario, this imbalance can be countered during selection
by preferring instances where the predicted class is underrepresented in the training
set. An instance is weighted by the following rule:
\begin{equation}
w_c = \frac{\text{\#instances} + \text{\#classes}}{\text{\#instances}_\text{c} + 1},
\end{equation}
where $c$ denotes the predicted class. We assume a symmetric Dirichlet prior with
$\alpha=1$, meaning that we have no prior knowledge of the class distribution,
and estimate the posterior after observing the total number of training instances as well as
the number of instances of class $c$ in the training set. The weight $w_c$ is then defined as
the inverse of the posterior to prefer underrepresented classes. It is multiplied with $v_{1vs2}(x)$ before aggregation
to obtain a final score.

\section{Experiments}
\label{sec:exp}
In the following, we present our evaluation.
First we show how the proposed aggregation metrics are able to enhance recognition performance while selecting new data for annotation.
After this, we will analyze the gained improvements when our proposed weighting scheme is applied.
\ifVISAPP
All code will be made available after publication.
\else
This paper describes work in progress. Code will be made available after conference publication.
\fi

\paragraph{Data}
We use the PASCAL VOC 2012 dataset~\cite{Everingham2010VOC} to assess the effects of our methods on learning.
To specifically measure incremental and active learning performance, both training
and validation set are split into parts A and B in two different random ways to obtain more general experimental results.
Part B is considered \enquote{new} and is comprised of images with the object classes
\texttt{bird}, \texttt{cow} and \texttt{sheep} (first way) or \texttt{tvmonitor}, \texttt{cat} and \texttt{boat} (second way). Part A contains all other 17 classes
and is used for initial training. The training set for part B contains 605 and 638 images for the first and second way, respectively.
The decision towards VOC in favor of recently published datasets was motivated by the conditions of the dataset itself.
Since it mainly contains images showing fewer objects, it is possible to split the data into a known and unknown part without having images containing classes from both parts of the splits.

\paragraph{Active Exploration Protocol}
Before an experimental run, the part B datasets are divided randomly into unlabeled batches of ten samples each.
This fixed assignment decreases the probability of
very similar images being selected for the same batch compared to always
selecting the highest valued samples, which would lead to less diverse batches.
This is valuable while dealing with data streams, \eg from camera traps, or data with low intra-class variance.
The construction of diverse unlabeled data batches is a well known topic in batch-mode active learning \cite{Settles2009ALS}.
However, the construction of diverse batches could lead to unintended side-effects and an evaluation of those is beyond the scope of the current study.
The unlabeled batch size is a trade-off between a tight feedback loop (smaller batches)
and computational efficiency (larger batches).
As side-effect of the fixed batch assignment, there are some samples left over during selection (\ie five for first way and eight for second way of splitting).

The unlabeled batches are assigned a value using the sum of the
active learning metric over all images in the corresponding batch as a meta-aggregation.
Other functions such as average or maximum could be considered too,
but are also beyond the scope of this paper.

The highest valued batch is selected
for an incremental training step \cite{Kaeding2016FDN}. The network
is updated using the annotations from the dataset in lieu of a human annotator.
Please note, annotations are not needed for update batch selection but for the update itself. This process
is repeated from the point of batch valuation until there are no unlabeled batches
left. The assignment of samples to unlabeled batches is not changed during an experimental run.

\paragraph{Evaluation}
\begin{algorithm*}
\caption{Detailed description of the experimental protocol. Please note that in an actual continuous learning scenario, new examples are
always added to $\mathfrak{U}$. The loop is never left because $\mathfrak{U}$ is never exhausted. The described splitting process would have
to be applied regularly.}
\label{alg:main}
\begin{algorithmic}
\Require{Known labeled samples $\mathfrak{L}$, unknown samples $\mathfrak{U}$, initial model $f_0$, active learning metric $v$}
\State
\State $\mathfrak{U} = \mathfrak{U}_1,\mathfrak{U}_2, \ldots \gets$ split of $\mathfrak{U}$ into random batches
\State $f \gets f_0$
\State
\While{$\mathfrak{U}$ is not empty}
  \State calculate scores for all batches in $\mathfrak{U}$ using $f$
  \State $\mathfrak{U}_{best} \gets$ highest scoring batch in $\mathfrak{U}$ according to $v$
  \State
  \State $\mathcal{Y}_{best} \gets$ annotations for $\mathfrak{U}_{best}$ \emph{human-machine interaction}
  \State $f \gets$ incrementally train $f$ using $\mathfrak{L}$ and $(\mathfrak{U}_{best}, \mathcal{Y}_{best})$
  \State
  \State $\mathfrak{U} \gets \mathfrak{U}\backslash\mathfrak{U}_{best}$
  \State $\mathfrak{L} \gets \mathfrak{L} \cup (\mathfrak{U}_{best}, \mathcal{Y}_{best})$
\EndWhile
\end{algorithmic}
\end{algorithm*}

We report mean average precision (mAP) as described in \cite{Everingham2010VOC} and validate each five new batches (\ie 50 new samples).
The result is averaged over five runs for each active learning metric and way of splitting for a total of ten runs.
As a baseline for comparison, we evaluate the performance of random selection, since there is no other work suitable for direct comparison without any adjustments as of yet.

\paragraph{Setup -- Object Detector}
We use YOLO as deep object detection framework \cite{Redmon2015YOLO}.
More precisely, we use the YOLO\hyp{}Small architecture as an alternative to larger object detection networks, because it allows for much faster training.
Our initial model is obtained by fine-tuning the \emph{Extraction} model\footnote{\url{http://pjreddie.com/media/files/extraction.weights}}
on part A of the VOC dataset for 24,000 iterations using the
Adam optimizer~\cite{Kingma2014Adam}, for each way of splitting the dataset into parts A and B, resulting in two initial models. The first half of initial training is completed
with a learning rate of 0.0001. The second half and all incremental experiments
use a lower learning rate of 0.00001 to prevent divergence.
Other hyperparameters match \cite{Redmon2015YOLO}, including the augmentation of training data using random crops, exposure or saturation adjustments.

\paragraph{Setup -- Incremental Learning}
Extending an existing CNN without sacrificing performance on known data is not a trivial task.
Fine\hyp{}tuning exclusively on new data leads to a severe degradation of performance on previously learned examples~\cite{Kirkpatrick2016Cat,Shmelkov2017Detector}.
We use a straightforward, but effective fine\hyp{}tuning method \cite{Kaeding2016FDN} to implement incremental learning.
With each gradient step, the mini-batch is constructed by randomly selecting from old and new examples with a certain probability of $\lambda$ or $1-\lambda$, respectively.
After completing the learning step, the new data is simply considered old data for the next step.
This method can balance known and unknown data performance successfully.
We use a value of 0.5 for $\lambda$ to make as few assumptions as possible
and perform 100 iterations per update. \Cref{alg:main} describes the protocol in more
detail. The method can be applied to YOLO object detection with some adjustments. Mainly,
the architecture needs to be changed when new classes are added. Because of the design of YOLO's
output layer, we rearrange the weights to fit new classes, adding 49 weights per class.

\subsection{Results}
We focus our analysis on the new, unknown data.
However, not losing performance on known data is also important.
We analyze the performance on the known part of the data (\ie part A of the VOC dataset) to validate our method.
In worst case, the mAP decreases from 36.7\% initially to 32.1\% averaged across all experimental runs and methods although three new classes were introduced.
We can see that the incremental learning method from \cite{Kaeding2016FDN}
causes only minimal losses on known data. These losses in performance are also
referred to as \enquote{catastrophic forgetting} in literature \cite{Kirkpatrick2016Cat}.
The method from \cite{Kaeding2016FDN} does not require additional model parameters or adjusted loss terms for added samples
like comparable approaches such as \cite{Shmelkov2017Detector} do, which is important for learning indefinitely.

Performance of active learning methods is usually evaluated by observing points on a learning curve (\ie performance over number of added samples).
In \cref{tbl:table}, we show the mAP for the new classes from part B of VOC at several intermediate learning steps as well as exhausting the unlabeled pool.
In addition we show the area under learning curve (AULC) to further improve comparability among the methods.
In our experiments, the number of samples added equals the number of images.

\paragraph{Quantitative Results -- Fast Exploration}
\begin{table*}[tb]
    \caption{
        Validation results on part B of the VOC data (\ie \emph{new classes only}).
        \textbf{Bold} face indicates block-wise best results, \ie best results with and without additional weighting ($\cdot + w$).
        \underline{Underlined} face highlights overall best results.
    }
    \centering
\begin{tabularx}{0.99\textwidth}{l | Y  Y  Y  Y  Y || Y }
    \hline
    & $~$50$~$ samples& 100 samples& 150 samples& 200 samples& 250 samples& All samples\\
    & mAP/AULC & mAP/AULC &  mAP/AULC & mAP/AULC & mAP/AULC & mAP/AULC \\
    \hline
    \multicolumn{7}{l}{\textbf{Baseline}}\\
    \hline
    \hspace{2mm}\emph{Random} & $8.7$ / $4.3$ & $12.4$ / $14.9$ & $15.5$ / $28.8$ & $18.7$ / $45.9$ & $21.9$ / $66.2$ & $32.4$ / $264.0$\\
    \hline
    \multicolumn{7}{l}{\textbf{Our Methods}}\\
    \hline
    \hspace{2mm}\emph{Max} & $\underline{\bf9.2}$ / $\underline{\bf4.6}$ & $12.9$ / $\bf15.7$ & $15.7$ / $30.0$ & $19.8$ / $47.8$ & $22.6$ / $69.0$ & $32.0$ / $\bf269.3$\\
    \hspace{2mm}\emph{Avg} & $9.0$ / $4.5$ & $12.4$ / $15.2$ & $15.8$ / $29.2$ & $19.3$ / $46.8$ & $\bf22.7$ / $67.8$ & $\underline{\bf33.3}$ / $266.4$\\
    \hspace{2mm}\emph{Sum} & $8.5$ / $4.2$ & $\underline{\bf14.3}$ / $15.6$ & $\bf17.3$ / $\underline{\bf31.4}$ & $\bf19.8$ / $\bf49.9$ & $22.7$ / $\bf71.2$ & $32.4$ / $268.2$\\
    \hline
    \hspace{2mm}\emph{Max}$~+~w$ & $\bf9.2$ / $\bf4.6$ & $13.0$ / $\underline{\bf15.7}$ & $17.0$ / $30.7$ & $20.6$ / $49.5$ & $23.2$ / $71.4$ & $\bf33.0$ / $271.0$\\
    \hspace{2mm}\emph{Avg}$~+~w$ & $8.7$ / $4.3$ & $12.5$ / $14.9$ & $16.6$ / $29.4$ & $19.9$ / $47.7$ & $22.4$ / $68.8$ & $32.7$ / $267.1$\\
    \hspace{2mm}\emph{Sum}$~+~w$ & $8.7$ / $4.4$ & $\bf13.7$ / $15.6$ & $\underline{\bf17.5}$ / $\bf31.2$ & $\underline{\bf20.9}$ / $\underline{\bf50.4}$ & $\underline{\bf24.3}$ / $\underline{\bf72.9}$ & $32.7$ / $\underline{\bf273.6}$\\
    \hline
\end{tabularx}

  \label{tbl:table}
\end{table*}

Gaining accuracy as fast as possible while minimizing the human supervision is one of the main goals of active learning.
Moreover, in continuous exploration scenarios, like faced in camera feeds or other continuous automatic measurements, it is assumed that new data is always available.
Hence, the pool of valuable examples will rarely be exhausted.
To assess the performance of our methods in this fast exploration context, we evaluate the models after learning learning small amounts of samples.
At this point there is still a large number of diverse samples for the methods to choose from,
which makes the following results much more relevant for practical applications than results on the full dataset.

In general, we can see that incremental learning works in the context of the new classes in part B of the data, meaning that we observe an improving performance for all methods.
After adding only $50$ samples, \emph{Max} and \emph{Avg} are performing better than passive selection while the \emph{Sum} metric is outperformed marginally.
When more and more samples are added (\ie $100$ to $250$ samples), we observe a superior performance of the \emph{Sum} aggregation.
But also the two other aggregation techniques are able to reach better rates than mere random selection.
We attribute the fast increase of performance for the \emph{Sum} metric to its tendency to select samples with many object inside which leads to more annotated bounding boxes.
However, the target application is a scenario where the amount of unlabeled data is huge or new data is approaching continuously and hence a complete evaluation by humans is infeasible.
Here, we consider the amount of images to be evaluated more critical as the time needed to draw single bounding boxes.
Another interesting fact is the almost equal performance of \emph{Max} and \emph{Avg} which can be explained as follows:
the VOC dataset consists mostly of images with only one object in them.
Therefore, both metrics lead to a similar score if objects are identified correctly.

We can also see that the proposed balance handling (\ie $\cdot + w$) causes slight losses in performance at very early stages up to $100$ samples.
At subsequent stages, it helps to gain noticeable improvements.
Especially for the \emph{Sum} method benefits from the sample weighting scheme.
A possible explanation for this behavior would be the following:
At early stages, the classifier has not seen many samples of each class and therefore suffers more from miss-classification errors.
Hence, the weighting scheme is not able to encourage the selection of rare class samples since the classifier decisions are still too unstable.
At later stages, this problem becomes less severe and the weighting scheme is much more helpful than in the beginning.
This could also explain the performance of \emph{Sum} in general.
Further details on learning pace are given later in a qualitative study on model evolution.
Additionally, the \emph{Sum} aggregation tends to select batches with many detections in it.
Hence, it is natural that the improvement is noticeable the most with this aggregation technique since it helps to find batches with many rare objects in it.

\paragraph{Quantitative Results -- All Available Samples}
In our case, active learning only affects the sequence of unlabeled batches if we train until there is no new data available.
Therefore, there are only very small differences between each method's results for mAP after training has completed.
The small differences indicate that the chosen incremental learning technique \cite{Kaeding2016FDN} is suitable for the faced scenario.
In continuous exploration, it is usually assumed that there will be more new unlabeled data available than can be processed.
Nevertheless, evaluating the long term performance of our metrics is important to detect possible deterioration over time compared to random selection.
In contrast to this, the differences in AULC arise from the improvements of the different methods during the experimental run
and therefore should be considered as distinctive feature implying the performance over the whole experiment.
Having this in mind, we can still see that \emph{Sum} performs best while the weighting generally leads to improvements.

\paragraph{Quantitative Results --- Class-wise Analysis}
\begin{figure}
    \centering
    \includegraphics[width=0.49\textwidth]{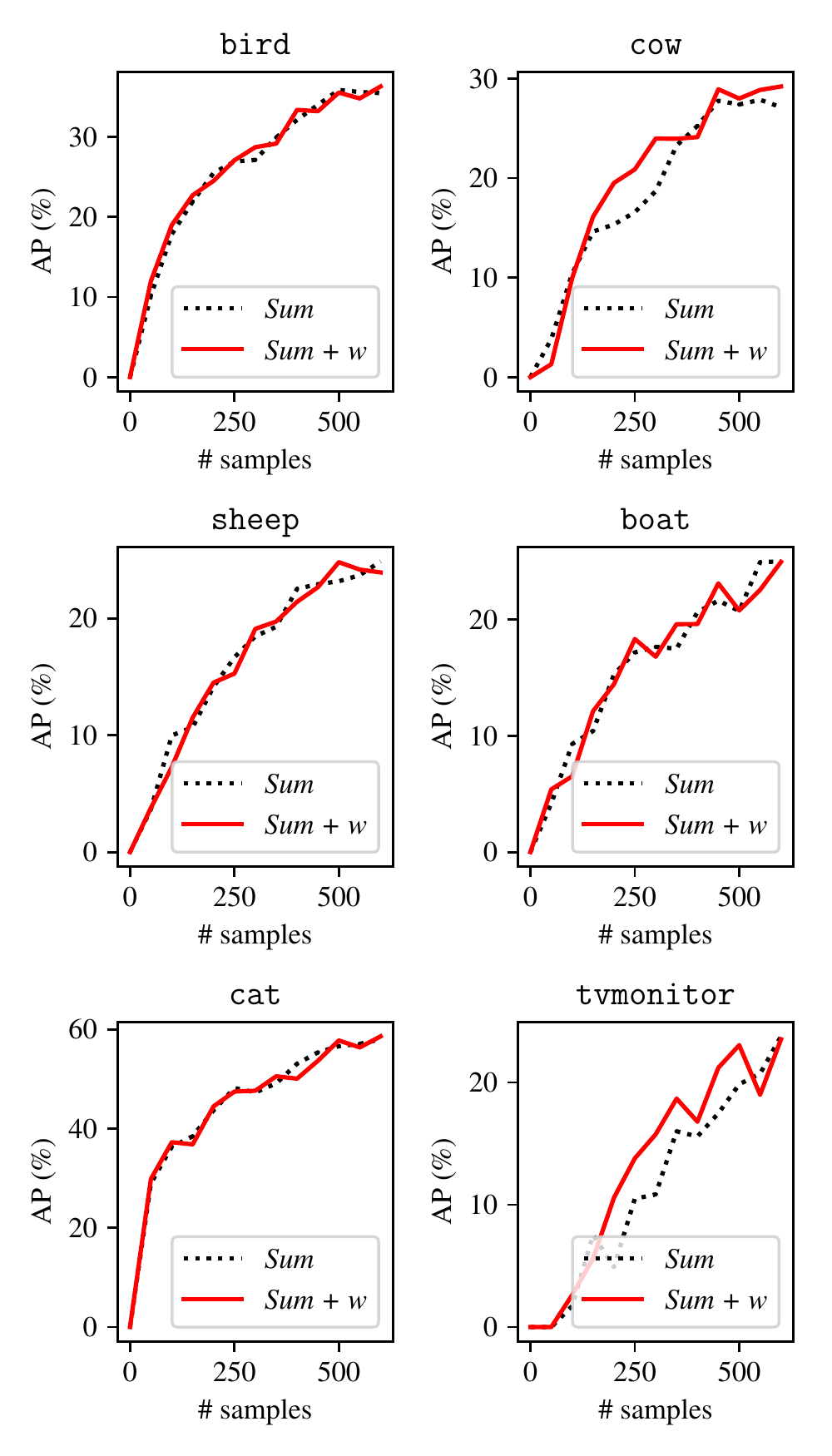}\\
    \includegraphics[width=0.49\textwidth]{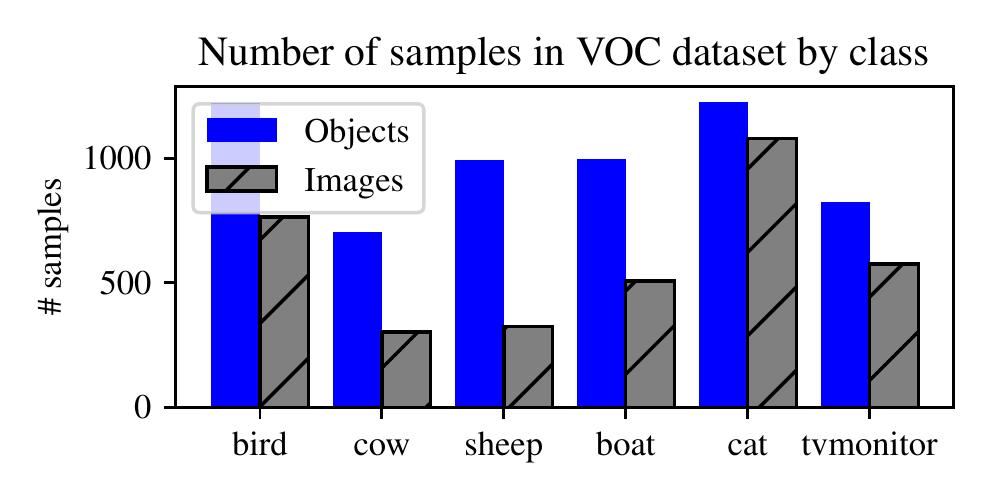}\\
    \caption{
        Class-wise validation results on part B of the VOC dataset (\ie, unknown classes).
        The first row details the first way of splitting (\texttt{bird},\texttt{cow},\texttt{sheep})
        while the second row shows the second way (\texttt{boat},\texttt{cat},\texttt{tvmonitor}).
        For reference, the distribution of samples (object instances as well as images with at least
        one instance) over the VOC dataset is provided in the third row.
    }
\label{fig:classwise}
\end{figure}
To validate the efficacy of our sample weighting strategy as discussed in \cref{sec:classwise},
it is important to measure not only overall performance, but to look at metrics for individual classes.
\cref{fig:classwise} shows the performance over time on the validation set for each individual class.
For reference, we also provide the class distribution over the relevant part of the VOC dataset,
indicated by number of object instances in total as well as number of images with at least one instance
in it.

In the first row, we observe an advantage for the weighted method when looking at the performance of \texttt{cow}.
Out of the three classes in this way of splitting \texttt{cow} has the fewest instances in the dataset.
The performance of \texttt{tvmonitor} in the second row shows a similar pattern, where it is also the class
with the lowest number of object instances in the dataset. Analyzing \texttt{bird} and \texttt{cat}, the top
classes by number of instances in each way of splitting, we observe only small differences in performance.
Thus, we can show evidence that our balancing scheme is able to improve performance on rare classes while it does not effect performance on frequent classes.

Intuitively, these observations are in line with our expectations regarding our handling of class imbalances,
where examples of rare classes should be preferred during selection. We start to observe the advantages after
around 100 training examples, because, for the selection to happen correctly, the prediction of the rare class needs to be
correct in the first place.

\paragraph{Qualitative Results -- Sample Valuation}
\newcommand{\mrowcell}[2][2cm]{\adjustbox{raise=5mm}{\pbox{17mm}{#2}}}
\newcommand{\bwwidth}{0.12\textwidth}
\newcommand{\bwheight}{12mm}
\newcommand{\bwvalign}{C}

\begin{figure*}
  \setlength\tabcolsep{1.5pt}
  \centering
  \begin{tabular}{L{17mm} c c c c c c c}
    \hline
    \multicolumn{8}{c}{\small Most valuable examples (highest score)}\\
    \hline
    \rule{0pt}{3.35\normalbaselineskip}
    \mrowcell{\small $Sum\:(+\:w)$} &
    \includegraphics[width=\bwwidth,height=\bwheight,valign=\bwvalign]{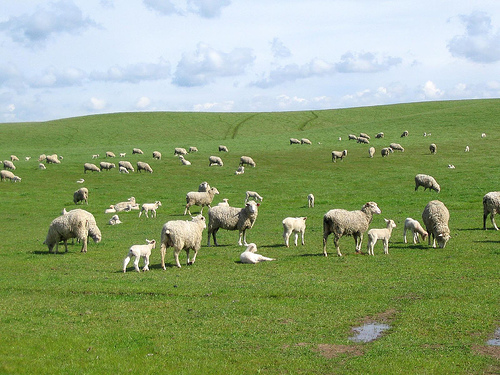} &
    \includegraphics[width=\bwwidth,height=\bwheight,valign=\bwvalign]{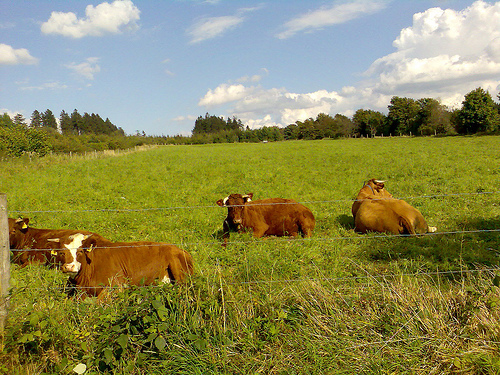} &
    \includegraphics[width=\bwwidth,height=\bwheight,valign=\bwvalign]{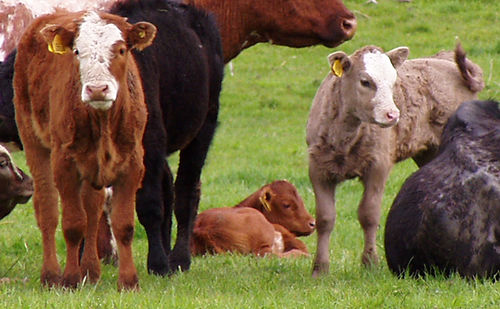} &
    \includegraphics[width=\bwwidth,height=\bwheight,valign=\bwvalign]{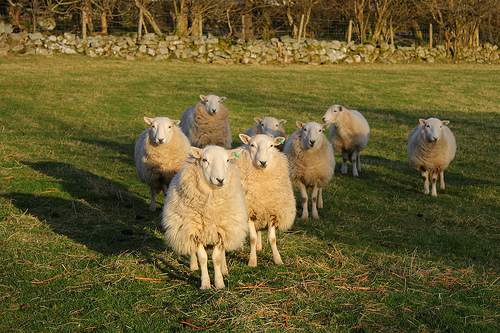} &
    \includegraphics[width=\bwwidth,height=\bwheight,valign=\bwvalign]{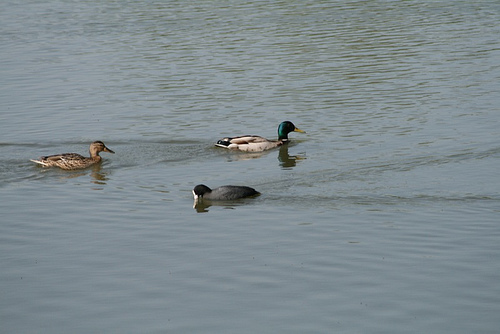} &
    \includegraphics[width=\bwwidth,height=\bwheight,valign=\bwvalign]{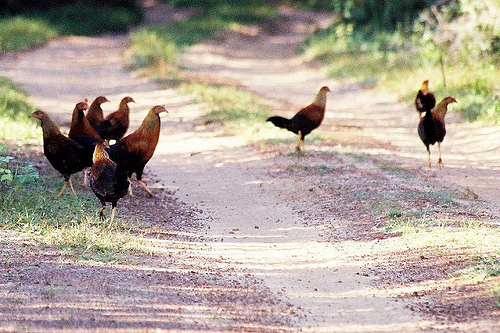} &
    \includegraphics[width=\bwwidth,height=\bwheight,valign=\bwvalign]{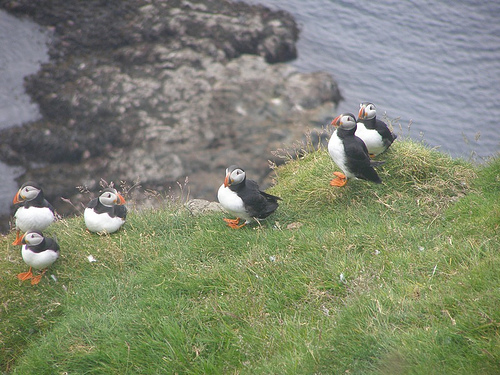} \\

    \mrowcell{\small $Avg\:(+\:w)$} &
    \includegraphics[width=\bwwidth,height=\bwheight,valign=\bwvalign]{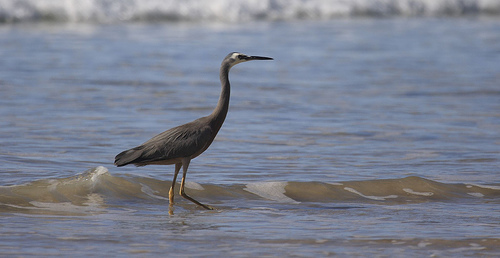} &
    \includegraphics[width=\bwwidth,height=\bwheight,valign=\bwvalign]{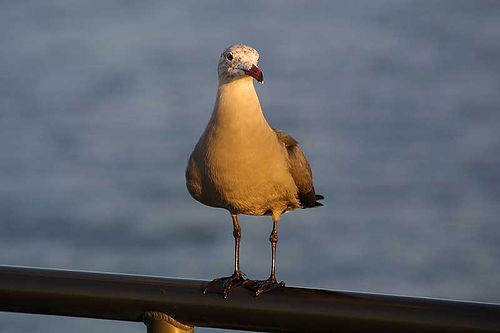} &
    \includegraphics[width=\bwwidth,height=\bwheight,valign=\bwvalign]{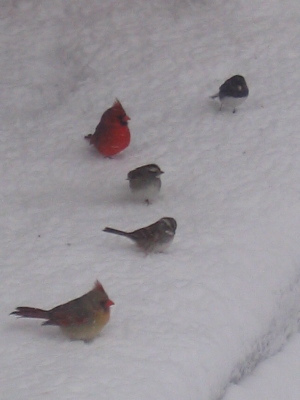} &
    \includegraphics[width=\bwwidth,height=\bwheight,valign=\bwvalign]{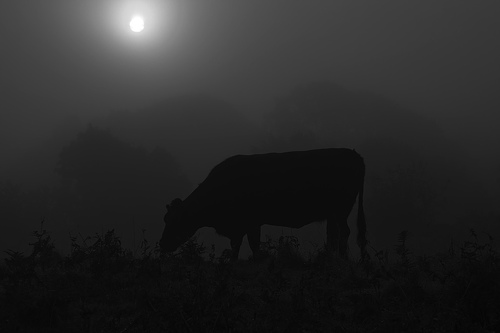} &
    \includegraphics[width=\bwwidth,height=\bwheight,valign=\bwvalign]{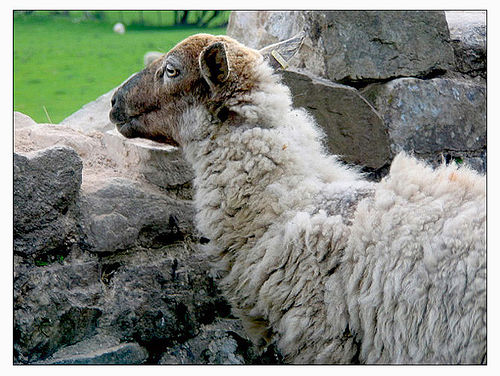} &
    \includegraphics[width=\bwwidth,height=\bwheight,valign=\bwvalign]{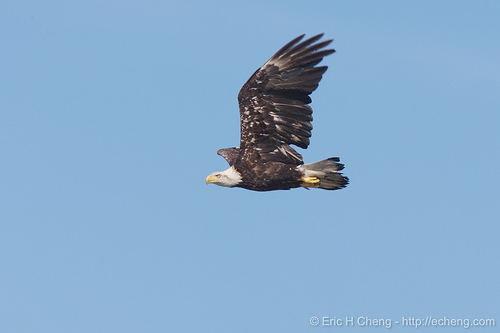} &
    \includegraphics[width=\bwwidth,height=\bwheight,valign=\bwvalign]{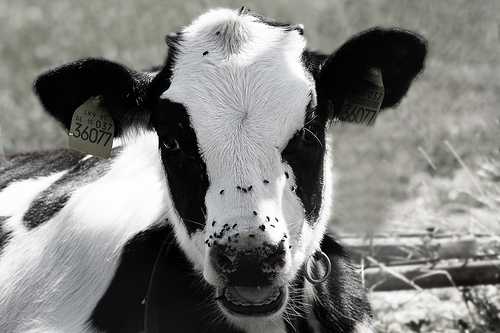} \\

    \mrowcell{\small $Max\:(+\:w)$} &
    \includegraphics[width=\bwwidth,height=\bwheight,valign=\bwvalign]{pictures/2008_008197.jpg} &
    \includegraphics[width=\bwwidth,height=\bwheight,valign=\bwvalign]{pictures/2010_004953.jpg} &
    \includegraphics[width=\bwwidth,height=\bwheight,valign=\bwvalign]{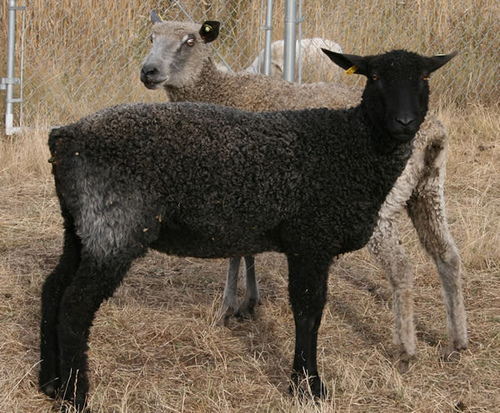} &
    \includegraphics[width=\bwwidth,height=\bwheight,valign=\bwvalign]{pictures/2008_008347.jpg} &
    \includegraphics[width=\bwwidth,height=\bwheight,valign=\bwvalign]{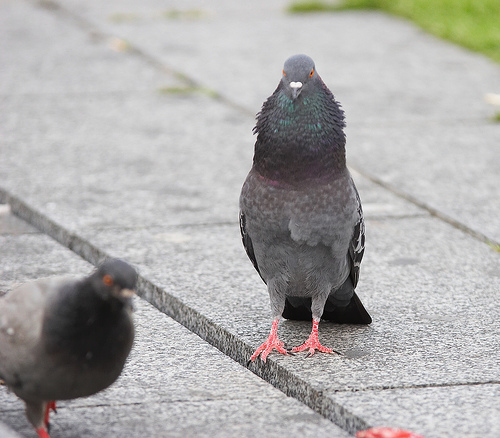} &
    \includegraphics[width=\bwwidth,height=\bwheight,valign=\bwvalign]{pictures/2010_003345.jpg} &
    \includegraphics[width=\bwwidth,height=\bwheight,valign=\bwvalign]{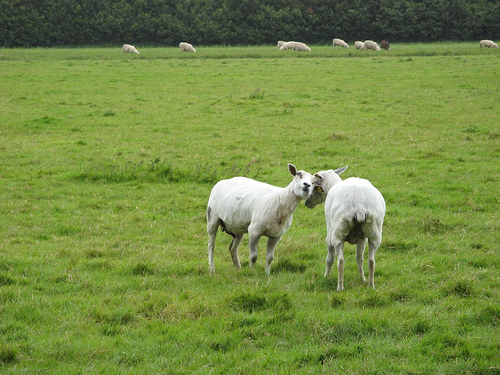} \\

    \hline
    \multicolumn{8}{c}{\small Least valuable examples (zero score)}\\
    \hline
    \rule{0pt}{3.35\normalbaselineskip}
    \mrowcell{\small All} &
    \includegraphics[width=\bwwidth,height=\bwheight,valign=\bwvalign]{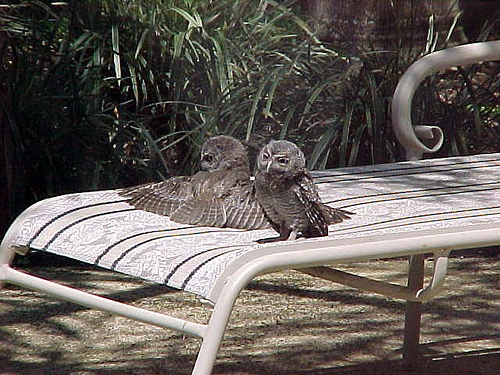} &
    \includegraphics[width=\bwwidth,height=\bwheight,valign=\bwvalign]{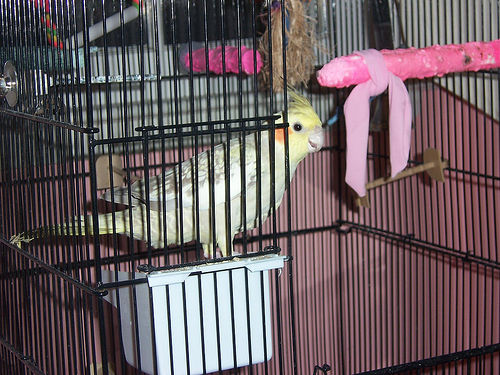} &
    \includegraphics[width=\bwwidth,height=\bwheight,valign=\bwvalign]{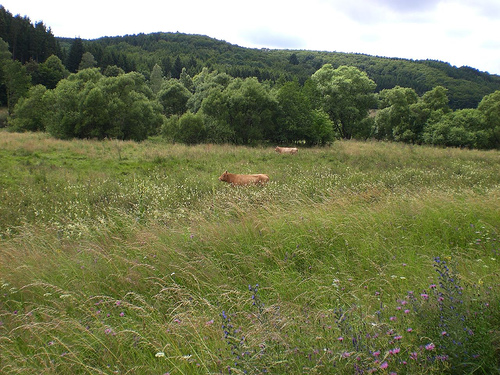} &
    \includegraphics[width=\bwwidth,height=\bwheight,valign=\bwvalign]{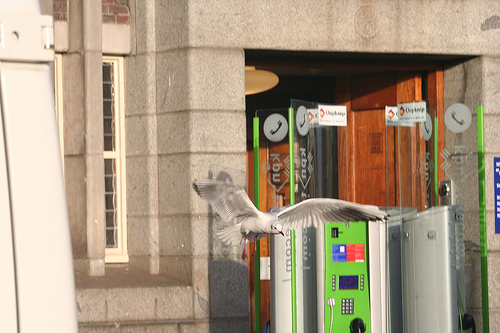} &
    \includegraphics[width=\bwwidth,height=\bwheight,valign=\bwvalign]{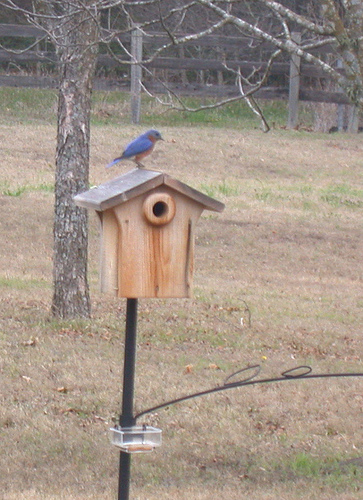} &
    \includegraphics[width=\bwwidth,height=\bwheight,valign=\bwvalign]{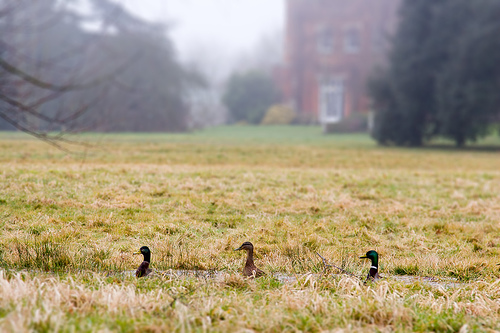} &
    \includegraphics[width=\bwwidth,height=\bwheight,valign=\bwvalign]{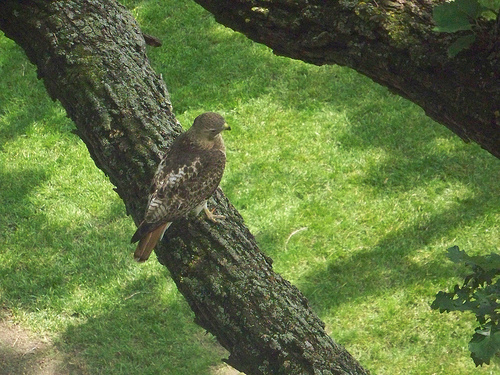}\\
    \hline
  \end{tabular}

    \vspace{3pt}
  \caption{Value of examples of \texttt{cow}, \texttt{sheep} and \texttt{bird}
    as determined by the \emph{Sum}, \emph{Avg} and
    \emph{Max} metrics using the \emph{initial} model. The top seven selection is not affected by using our weighting method to counter
    training set class imbalaces.}
  \label{fig:bestworst}
\end{figure*}

We calculate whole image scores over \texttt{bird}, \texttt{cow} and \texttt{sheep} samples
using our corresponding \emph{initial} model trained on the remaining classes for the first way of splitting.
\Cref{fig:bestworst} shows those images that the three aggregation
metrics consider the most valuable. Additionally, common zero scoring images
are shown. The least valuable
images shown here are representative
of all proposed metrics because they do not lead to any detections using the
\emph{initial} model.
Note that there are more than
seven images with zero score in the training dataset.
The images shown in the figure have been selected randomly.

Intuitively, the \emph{Sum} metric should
prefer images with many objects in them over single objects, even if
individual detection values are low. Although VOC contains mostly of images with a single object, all seven of the highest scoring
images contain at least three objects.
The \emph{Average} and \emph{Maximum} metric prefer almost identical images since the average and maximum are used to be nearly equal for few detections.
With few exceptions, the most valuable images contain pristine examples
of each object. They are well lit and isolated. The objects in the zero scoring
images are more noisy and hard to identify even for the human viewer,
resulting in fewer reliable detections.

\paragraph{Qualitative Results -- Model Evolution}
\newcommand{\mrowcellev}[2][5cm]{\adjustbox{raise=6mm}{\pbox{25mm}{\centering #2}}}
\newcommand{\evwidth}{0.11\textwidth}
\newcommand{\evheight}{12mm}
\newcommand{\evvalign}{C}

\begin{figure*}[tb]
    \setlength\tabcolsep{1.5pt}
    \centering
    \scriptsize
    \begin{tabular}{cccccc}
        \hline
        & \multicolumn{3}{c}{New classes (part B)} & \multicolumn{2}{c}{Known classes (part A)}\\
        & \texttt{bird} & \texttt{cow} & \texttt{sheep} & \texttt{aeroplane} & \texttt{car}\\
        \hline
        \rule{0pt}{4\normalbaselineskip}
        \mrowcellev{Initial prediction} &
        \includegraphics[width=\evwidth,height=\evheight,valign=\evvalign]{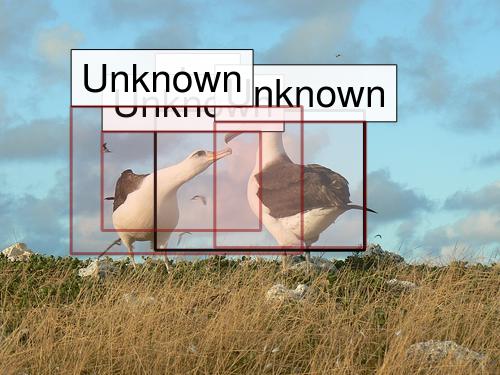} &
        \includegraphics[width=\evwidth,height=\evheight,valign=\evvalign]{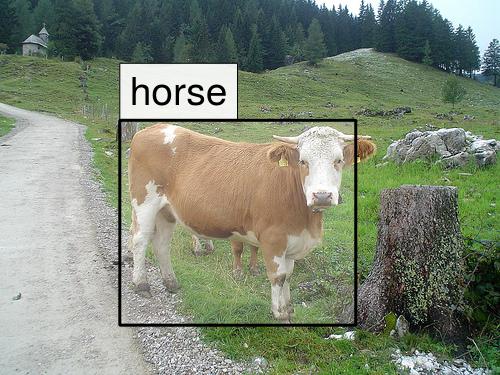} &
        \includegraphics[width=\evwidth,height=\evheight,valign=\evvalign]{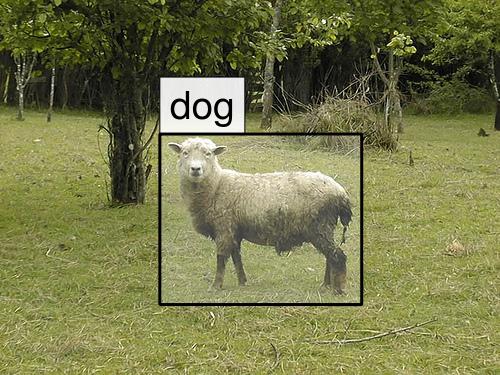} &
        \includegraphics[width=\evwidth,height=\evheight,valign=\evvalign]{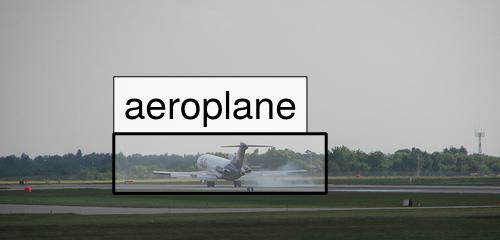} &
        \includegraphics[width=\evwidth,height=\evheight,valign=\evvalign]{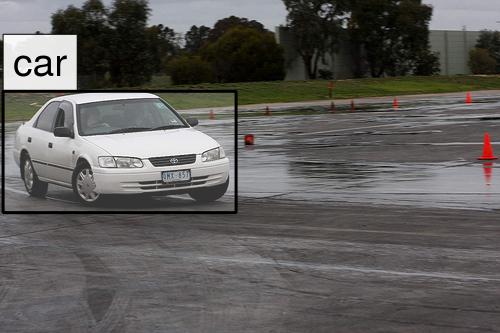}\\
        \mrowcellev{After 50 samples} &
        \includegraphics[width=\evwidth,height=\evheight,valign=\evvalign]{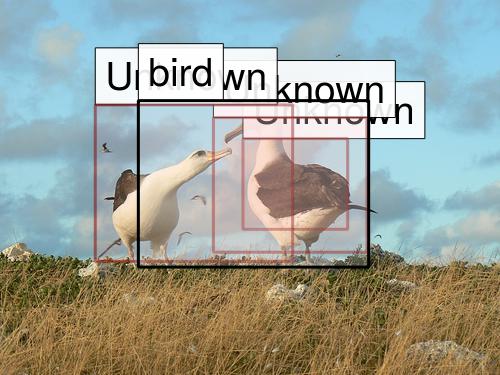} &
        \includegraphics[width=\evwidth,height=\evheight,valign=\evvalign]{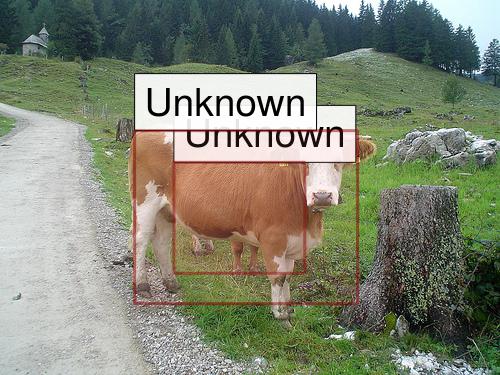} &
        \includegraphics[width=\evwidth,height=\evheight,valign=\evvalign]{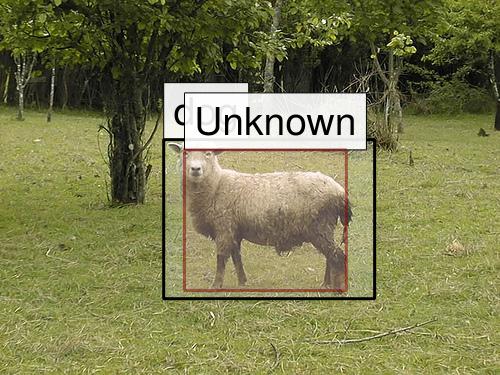} &
        \includegraphics[width=\evwidth,height=\evheight,valign=\evvalign]{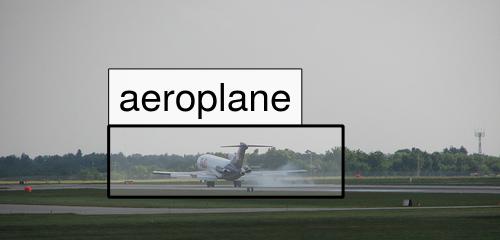} &
        \includegraphics[width=\evwidth,height=\evheight,valign=\evvalign]{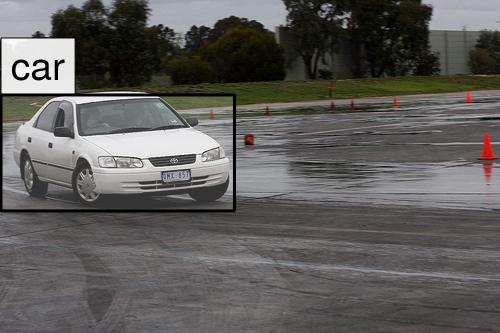}\\
        \mrowcellev{After 150 samples} &
        \includegraphics[width=\evwidth,height=\evheight,valign=\evvalign]{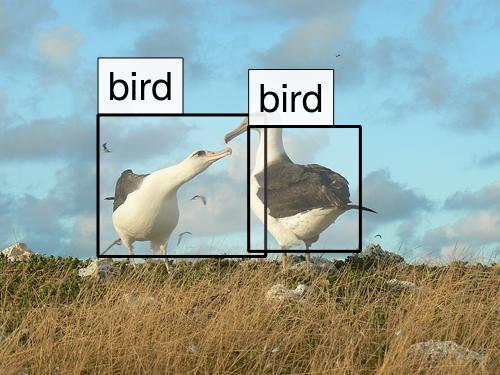} &
        \includegraphics[width=\evwidth,height=\evheight,valign=\evvalign]{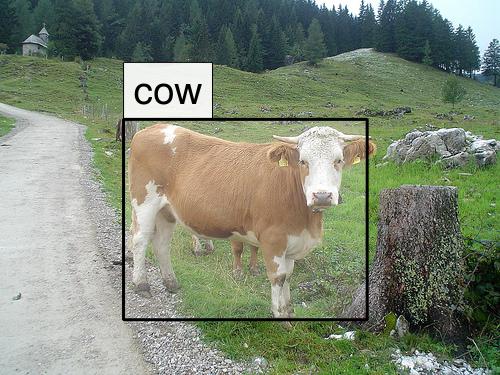} &
        \includegraphics[width=\evwidth,height=\evheight,valign=\evvalign]{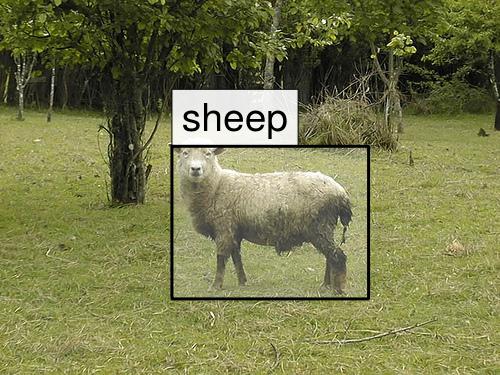} &
        \includegraphics[width=\evwidth,height=\evheight,valign=\evvalign]{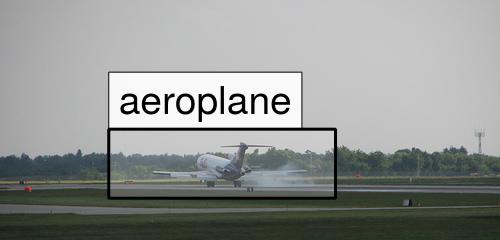} &
        \includegraphics[width=\evwidth,height=\evheight,valign=\evvalign]{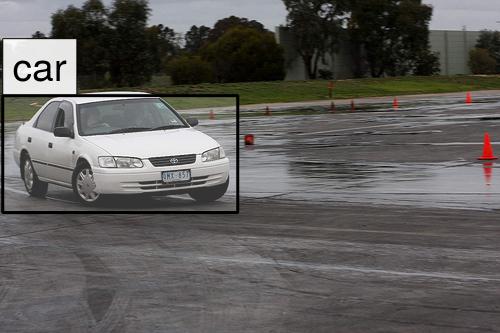}\\
        \hline
    \end{tabular}
    \vspace{3pt}
    \caption{Evolution of detections on examples from validation set.}
    \label{fig:evolution}
\end{figure*}

Observing the change in model output as new data is learned
can help estimate the number of samples needed
to learn new classes and identify possible confusions. \cref{fig:evolution}
shows the evolution from initial guesses to correct detections after learning
150 samples, corresponding to an fast exploration scenario. For selection,
the \emph{Sum} metric is used.

The class confusions shown in the figure are typical for this scenario.
\texttt{cow} and \texttt{sheep} are recognized as visually similar \texttt{dog}, \texttt{horse} and \texttt{cat}.
\texttt{bird} is often classified as \texttt{aeroplane}.
After selecting and learning 150 samples,
the objects are detected and classified correctly and reliably.

During the learning process, there are also \emph{unknown} objects.
Please note, being able to mark objects as \emph{unknown} is a direct consequence of using YOLO.
Those objects have a detection confidence above the required threshold, but no classification is certain enough.
This property of YOLO is important for the discovery of objects of new classes.
Nevertheless, if similar information is available from other detection methods, our techniques could easily be applied.

\section{Conclusions}
\label{sec:conclusion}
In this paper, we propose several uncertainty\hyp{}based active learning metrics for object detection.
They only require a distribution of classification scores per
detection. Depending on the specific task, an object detector that will
report objects of unknown classes is also important.
Additionally, we propose a sample weighting scheme to balance selections among classes.

We evaluate the proposed metrics on the PASCAL VOC 2012 dataset \cite{Everingham2010VOC} and offer
quantitative and qualitative results and analysis.
We show that the proposed metrics are able to guide the annotation process efficiently which leads to superior performance in comparison to a random selection baseline.
In our experimental evaluation, the \emph{Sum} metric is able to achieve best results overall which can be attributed to the fact that it tends to select batches with many single objects in it.
However,
the targeted scenario is an application with huge amounts of unlabeled data where we consider the amount of images to be evaluated as more critical than the time needed to draw single bounding boxes.
Examples would be camera streams or camera trap data.
To expedite annotation, our approach could be combined with a weakly supervised learning approach as presented in \cite{Papdopoulos2016Semi}.
We also showed that our weighting scheme leads to even increased accuracies.

All presented metrics could be applied to other deep object detectors, such as the variants of SSD \cite{Liu2015SSD},
the improved R-CNNs \eg \cite{Ren2015FasterRCNN} or the newer versions of YOLO \cite{Redmon2016YOLOv2}.
Moreover, our proposed metrics are not restricted to deep object detection and could be applied to arbitrary object detection methods if they fulfill the requirements.
It only requires a complete distribution of classifications scores per detection.
Also the underlying uncertainty measure could be replaced with arbitrary active learning metrics to be aggregated afterwards.
Depending on the specific task, an object detector that will report objects of unknown classes is also important.

The proposed aggregation strategies also generalize to selection of images based on segmentation results or any other type of image partition.
The resulting scores could also be applied in a novelty detection scenario.


\bibliographystyle{apalike}
\bibliography{paper}

\end{document}